# ADMEOOD: Out-of-Distribution Benchmark for Drug Property Prediction

Shuoying Wei[1,2,3], Xinlong Wen[1,2,3], Songquan Li[1,2,3], Lida Zhu[1,2,3], Rongbo Zhu[1,2,3]

1 College of Informatics, Huazhong Agricultural University, 430070, Wuhan, China
2 Shenzhen Institute of Nutrition and Health, Huazhong Agricultural University, 518000, Shenzhen, China
3 Shenzhen Branch, Guangdong Laboratory for Lingnan Modern Agriculture, Genome Analysis Laboratory of the Ministry of Agriculture, and Agricultural Genomics Institute at Shenzhen, Chinese Academy of Agricultural Sciences, 518000, Shenzhen, China

## Abstract

Obtaining accurate and valid information for drug molecules is a crucial and challenging task. However, chemical knowledge and information have been accumulated over the past 100 years from various regions, laboratories, and experimental purposes. Little has been explored in terms of the out-of-distribution (OOD) problem with noise and inconsistency, which may lead to weak robustness and unsatisfied performance. This study proposes a novel benchmark ADMEOOD, a systematic OOD dataset curator and benchmark specifically designed for drug property prediction. ADMEOOD obtained 27 ADME (Absorption, Distribution, Metabolism, Excretion) drug properties from Chembl and relevant literature. Additionally, it includes two kinds of OOD data shifts: Noise Shift and Concept Conflict Drift (CCD). Noise Shift responds to the noise level by categorizing the environment into different confidence levels. On the other hand, CCD describes the data which has inconsistent label among the original data. Finally, it tested on a variety of domain generalization models, and the experimental results demonstrate the effectiveness of the proposed partition method in ADMEOOD: ADMEOOD demonstrates a significant difference performance between in-distribution and out-of-distribution data. Moreover, ERM (Empirical Risk Minimization) and other models exhibit distinct trends in performance across different domains and measurement types.

## Introduction

The traditional drug discovery process is extremely time-consuming and expensive. Especially, up to 50% of clinical trial failures have been attributed to deficiencies in ADMET (Absorption, Distribution, Metabolism, Excretion, Toxicity) properties (Kennedy 1997; Kola and Landis 2004). To address this shortcoming, methods with maximum efficiency and cost-effectiveness are urgently needed. In recent years, the pharmaceutical industry has been collecting experimental ADME properties of numerous compounds and utilizing them to forecast the ADME properties of novel compounds. The widespread use of artificial intelligence techniques such as deep learning has led to a dramatic improvement in the drug property prediction. These experimental ADME properties are based on chemical knowledge and information that have been accumulated over the past 100 years from various regions, laboratories, and experimental purposes. This introduces out-of-distribution (OOD) problem with noise and inconsistency, resulting in weak model robustness and suboptimal performance, such as ChEMBL, whereas the real bioassay data has various factors including but not limited to different confidence levels for activities measured through experiments, unit-translation errors, repeated citations of single measurements and different noise (Kramer et al. 2012; Cortés-Ciriano and Bender 2016), etc.

Existing machine learning are mostly based on an underlying hypothesis that training and testing data are independently sampled from and identical environment, yet real-world environments are often noise and inconsistent, which requires the model to effectively handle distribution shifts. There are two commonly studied data distribution shifts: Noise Shift and Concept Conflict Drift (CCD). Noise Shift refers to the presence of random or non-random noise in the data, which may be errors, disturbances, or other undesired variations that affect the performance of the model. It responds to the noise level by categorizing the environment into different confidence levels and segregating the data accordingly (Ji et al. 2023). CCD refers to the situation where the labels of the data appear to conflict with each other due to different sources of data and experimental scenarios, which may lead to degradation of model generalization performance or unstable prediction results. CCD describes the data which has inconsistent label among the original data. To address OOD problem substantially, sev-eral benchmarks have been curated (Gui et al. 2022; Ji et al. 2023) to evaluate different algorithms (Ganin et al. 2016; Arjovsky et al. 2019). However, existing benchmarks lack in several aspects, as detailed in Related Work. It is crucial to determine their boundary conditions.

To help accelerate research by simplifying systematic comparisons between data collection and implementation method, a systematic OOD dataset curator and benchmark specifically designed for drug property prediction called ADMEOOD is presented. The benchmark automates data management and OOD benchmark testing, and can serve as a testing platform in the field of drug property prediction to evaluate the performance of models on OOD data. In this study, ADMEOOD provides two different methods of OOD data partitioning, contains more ADME drug property information, as well as different domains, noise levels, and measurement types. The following summarizes the main contributions of this article:

- **Automated dataset curator:** Providing an automated and customizable pipeline for managing OOD datasets for drug property prediction from the large-scale bioassay deposition site ChEMBL, and filtering the 27 drug ADME properties needed from relevant references.
- **Domain annotations:** Two approaches are provided to generate specific domains that match biochemistry domain knowledge and to carefully design data environments.
- **Data Shift Method:** Noise Shift and Concept Conflict Drift (CCD). Noise Shift responds to the noise level by categorizing the environment into different confidence levels. On the other hand, CCD describes the data which has inconsistent label among the original data. And built a set of OOD datasets based on this.
- **Rigorous OOD benchmarking:** This benchmark tests four contemporary classic OOD generalization algorithms on 24 realized data set instances. It also compares and analyzes the results of these experiments, and gain insight into OOD study under noise for drug property prediction.

## Related Work

In this section, we review related literature from the perspectives of drug property prediction, out-of-distribution generalization, and commonly used drug properties benchmarks.

**Drug Property Prediction.** Accurate prediction of drug molecular properties plays a crucial role in drug discovery. Traditional feature engineering-based methods rely on handcrafted descriptors or fingerprints, which requires a significant amount of human expert knowledge (Sheridan et al. 2016; Gertrudes et al. 2012). In recent years, the rapid advancement of deep learning methods has provided a data-driven approach that can automatically learn molecular representations from primary data in the end-to-end training (Wieder et al. 2020). However, the existing models heavily depend on the quality of data, and lack a fine-grained annotation of distribution shifts, result in severe performance degradation. Therefore, it is urgent to establish a comprehensive dataset tailored for drug property prediction to address the OOD generalization problem.

**Drug Properties Benchmark.** Chemical knowledge and information have been accumulated over the past 100 years from various regions, laboratories, and experimental purposes. ChEMBL (Gaulton et al. 2011) is a large-scale bioanalytical database that gathers data from biology and pharmacology experiments from all over the world that aims to capture medicinal chemistry data and knowledge across the pharmaceutical R&D process (Ji et al. 2023). HIV is a small-scale real-world molecular dataset adapted from MoleculeNet which contains property data for over 700,000 compounds (Wu et al. 2018). The SIDER is a public database containing 1427 approved drugs and their adverse drug reactions (Kuhn et al. 2016). In SIDER, the drug side effects are grouped into 27 systemic organ classes according to the MedDRA classifications. These databases provide a great amount of testing data. Due to the absence of criteria to analysis data with OOD shifts such as noise and inconsistency, the training and evaluation subsets are often randomly divided, which can lead to overly optimistic model evaluations. Hence, rational annotations for OOD data are crucial and required.

**OOD Generalization methods.** In order to improve the robustness of models against distribution shifts, existing domain generalization algorithms tend to learn invariant representations that can generalize across domains. There are two main methods: domain alignment and invariant prediction. Domain alignment aims to minimize the divergence of feature distributions between different domains in distance measurement. The minimization of feature discrepancy can be conducted over various distance metrics, including Wasserstein distance (Zhou et al. 2021). Another research is established to generate new samples or domains through data augmentation to enhance the consistency of feature representations, such as Mixup (Xu et al. 2020). For learning invariant predictors, the main concept is to improve the correlation between invariant representations and labels (Ji et al. 2023), including Invariant Risk Minimization (Arjovsky et al. 2019) and DeepCORAL (Sun and Saenko 2016). In our study, ADMEOOD benchmark provides comprehensive performance assessment for over four state-of-art OOD generalization algorithms.

**OOD Benchmark.** Among the existing OOD benchmark tests, most of the benchmarks are not specifically designed for the drug property prediction domain. For example，DrugOOD (Ji et al. 2023) focuses on domain generalization for molecules, and gains insight into OOD learning under noise for AI-aided drug discovery. WILDS (Koh et al. 2021) benchmark was proposed to reflect various levels of distribution shifts that may occur in real-world scenarios, including covering shifts across cameras for wildlife monitoring, hospitals for tumor identification, users for product rating estimation and so on. In terms of graph OOD, the datasets in GOOD (Gui et al. 2022) are not all real-world data, but mostly synthetic and semi-synthetic. ADMEOOD is a benchmark designed specifically for the field of drug property prediction, and the data in ADMEOOD are from real experimental records. In addition, ADMEOOD proposes new dataset segmentation methods based on the characteristics associated with drug properties data, and compares between each segmentation method, resulting in a more comprehensive benchmark.

## ADMEOOD

In this study, a comprehensive benchmark for facilitating OOD research in drug property predic-tion: ADMEOOD is presented, which based on ADME properties in large-scale bioassay databases ChEMBL and related study.

The overall architecture of the ADMEOOD benchmark is shown in Figure 1, and three parts of the benchmark are presented separately next.

### Automated Data Curator

Before constructing the OOD dataset, the basic data are subjected to the normalization processes in DrugOOD. The specifics details of the treatments are demonstrated in Figure 2, including noise level annotations, uncertainty value data setting, filtering of the drug property data, and defining and delineating the data domains.

**Noise level setting.** Bioactivity data exhibits a multitude of diverse noise sources. How to learn under noise is an important issue and should be delicate handled. Inspired by DrugOOD, which summarize the various types of noise and delineate different subsets based on noise severity, ADMEOOD combined disordered data with two different levels of noise using various filter configurations. Specific division details are presented in Appendix A.

**Filtering drug property data, handling uncertain values, and removing empty values.** When filtering and processing drug property data, it is necessary to identify some indicators, namely ADME characteristics, that are most relevant to drug property prediction from the literature on drug property prediction. After the screening process, various properties of 27 drug compounds to assess their absorption, distribution, metabolism and elimination are obtained. These properties include the partition coefficient (logp) (Xing and Glen 2002; Eros et al. 2002; Benfenati et al. 2003), the molecular weight of free radicals of the compounds (MW_FREEBASE) (Riedel et al. 2013), the activity value of the compound molecule (pchembl_value) (Burggraaff et al. 2020; Tan et ter automatic data curator, we collected 24 ADMEOOD datasets. Then the generated dataset is loaded to provide a comprehensive testbed for various types of modules, such as algorithms and network frameworks in a flexible manner.

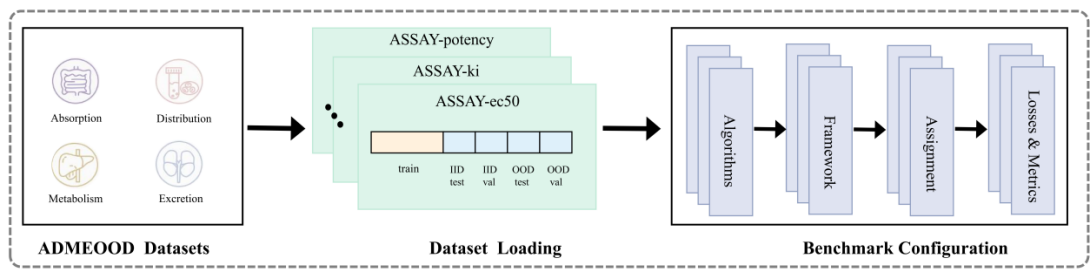

Figure 1: Overview of the ADMEOOD benchmark process. The ADMEOOD gathers drug characteristics, including absorption, distribution, metabolism and excretion to eatablish a comprehensive benchmark. Af-

al. 2022), acid-base dissociation constants (pka) (Cruciani et al. 2009; Alexov et al. 2011; Rupp et al. 2011; Witham et al. 2011), SMILES sequences of compounds (Wang et al. 2019), and drug toxicity records (Greene et al. 2002; Mayr et al. 2016). The specific criteria for each property can be found in Appendix B. When there is missing data in these property records, this record will be deleted from the dataset to ensure the quality and accuracy of the basic data.

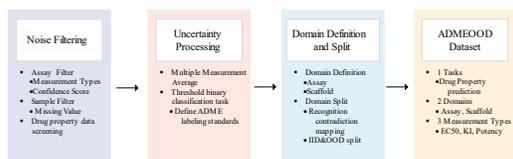

Figure 2: Normalization process of the drug properties prediction basis data.

**Drug property Data Label Definition.** In addition to filtering out the relevant metrics mentioned earlier in the literature on drug property prediction, The effect of the values taken for these metrics on the properties and pharmacological activities of the drugs is analyzed. For example, the value of the acid-base dissociation constant of a drug can affect its ionization state and solubility at different pH values. When the acid-base dissociation constant value of a drug is similar to the pH of the environment, the drug is more likely to be in an ionized state, which results in higher solubility. In contrast, when the value of the acid-base dissociation constant of a drug differs significantly from the pH of the environment, the drug is more likely to exist in a nonionic form and may have a lower solubility. The solubility of a drug directly affects its dissolution rate, absorption properties and bioavailability. The process and reasoning behind the identification of drug property data labels are explained in Figure 3.

## OOD dataset design

In this section, two methods for establishing OOD datasets for drug properties with bias: Noise Shift and CCD are introduced.

**Noise Shift.** In environments with varying levels of noise, there were changes in how the data samples were labeled, resulting in inconsistent labeling of the same sample in different noisy environments. Meanwhile, the increasing noise level brings more data. This leads to changes in the distribution of data, which in turn affects the model's prediction and classification results for new samples.

In an environment with low noise levels, data quality is better. In addition, the confidence level of the labels is relatively high. The accuracy of the model in predicting results will also be higher. In high-level noise environments, the data quality is low, hence the accuracy of the model's prediction results may be affected. A range of distributed data by dividing them into different confidence environments is reflected. The method used to obtain out-of-distribution data is called the confidence environment shift method.

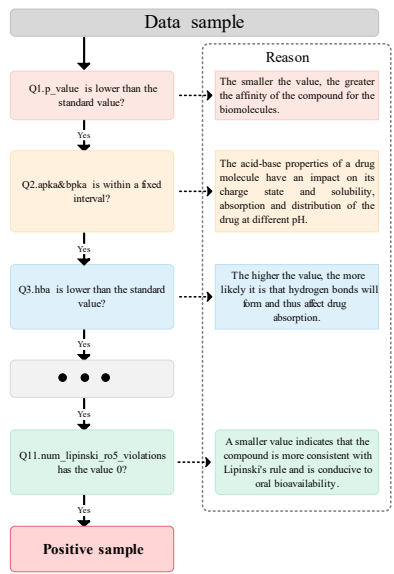

Figure 3: Flowchart of drug property data label definition. This figure shows the 11 assessment indicators for drug compound records, which will be marked as positive samples if all are met.

**Concept Conflict Drift (CCD).** There are a kind of label conflicts exists in multiple experimental records, that one drug has inconsistent labels in different experiments. This kind of label conflict describes a sample has incorrect or inconsistent labels. In general, the relationshipbetween the training sampleand labelis single-valued, as indicated by formula:

$$\forall x \in X \, \exists \, ! \, y \in Y \big( (x, y) \in f \big) \qquad (1)$$

where the relation f becomes a mapping from $X$ to $Y$, denoted as $f: X \to Y, x \to y = f(x)$.

But in some cases, the same data may be labeled as label A in one experiment, with the mapping rela-tionship: $f: X^{train} \rightarrow Y, x \rightarrow y_1 = f(x)$; while in another experiment, it may be labeled as label B, with the mapping relationship:
$f': X^{test} \rightarrow Y, x \rightarrow y_2 = f'(x)$.
This "one-to-many" relationship does not satisfy the unique determinism in the mapping. As a result, samples can be misclassified into multiple categories, which leads to inaccurate predictions. The prediction classification task in Figure 4 distinguishes two domains with yellow and green colors. Additionally, five scenarios are divided based on the distribution of sample data in different situations. In the Normal section, each data sample is uniquely identified by a category label within a domain. Anomaly shows anomalies in real-world data: a single drug compound sample is labeled with multiple categories within the same field. Data with conflicting tags is defined as shift data. By identifying the data with distribution shift through the contradictions between the data labels, A benchmark dataset for drug property prediction

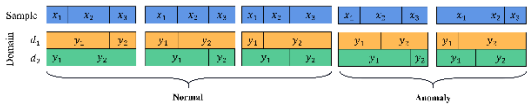

is constructed. The method that divides the dataset based on contradictions is called CCD.

Figure 4: Illustration of Conflicts Mapping Shift. Each concept includes two domains, and each domain has spurious correlation with a apecific output in a concept. For example, in the first sample of Normal, $x_2$ in the orange domain is highly correlated with $y_2$, but the $x_2$ in the orange domain of Anomaly for this sample is correlated with both $y_1$ and $y_2$. There is a conflict.

## Data loading

The self-constructed OOD benchmark dataset contains data on compound molecules, which are represented by their own sequence format called SMILES (Weininger et al. 1988). Therefore, it can be processed using a model that handles sequential data in a straightforward manner. But this one-dimensional linearization of the molecular structure is highly dependent on the traversal order of the molecular graph. This means that two atoms that are close in sequence may be far apart or even unrelated in the actual 2D/3D structure. The specific format of these sequences may contribute to the accuracy of the model's prediction results, therefore, the data is converted into molecular maps as input to the model during the loading process. This measure is taken to eliminate information loss in sequence models. Compared to tra-ditional molecular representations, molecular maps can provide more comprehensive structural information, allowing for more accurate characterization of molecules. Molecular graphs also utilize graph neural networks (GNN) to propagate and aggregate information, thereby generating embedded representations of chemical structure information. This effectively allows for the learning of feature representations of molecules, including node features and edge features, and improves the performance of drug property prediction models.

## Benchmark test

After loading the benchmark dataset, various modules, such as algorithms and network architectures, are utilized to evaluate the benchmark for drug property prediction. Based on this, more algorithms for OOD generalization can be developed flexibly.

**Domain generalization algorithm.** The Empirical Risk Minimization (ERM) algorithm is a classical machine learning algorithm used to solve domain adaptation and domain generalization problems. The basic idea of this algorithm is to train the model by minimizing the empirical risk. IRM algorithm is a novel learning approach that aims to estimate nonlinear, invariant, causal predictors from diverse training environments in order to achieve robust generalization (Arjovsky et al. 2019). DeepCORAL refers to an extension of CORAL that learns a nonlinear transformation to calibrate the correlation of activations in the middle layers of the network (Sun and Saenko 2016). On the basis of the ERM algorithm, Mixup proposes a new data augmentation method that utilizes linear interpolation to generate augmented data (Zhang et al. 2017).

**Model backbone network.** The expressive power of the model largely depends on the network structure. How to select a suitable backbone network, i. e., with better objective function fitting ability and robustness to noise, is a hot research topic in the problem of non-distribution of data (Ji et al. 2023). This study benchmarks and evaluates some of the following graph-based backbone networks: the GCN (Kipf and Welling 2016), the GIN (Xu et al. 2018), the GAT (Velickovic et al. 2017) and the MGCN (Lu et al. 2019). The reason for choosing graph-based networks is that they establish representation models by learning the sequential structure information of molecules (Zhu et al. 2021). Compared to traditional methods of molecular representation, this approach utilizes end-to-e-nd learning, eliminating the need for manual design of molecular descriptors.

The correlation network transforms the various feature representations of the compounds into graphical structures that are utilized as inputs to the model. As shown in Figure 5, we investigated three feature representations: the SMILES sequence of the compound molecule, the molecular scaffolding representation, and the molecular map representation. Molecular processing tools, such as RDKit (Landrum 2013) or Open Babel (O'Boyle et al. 2011), are utilized to convert molecules into molecular entities. These objects are then converted into undirected graphical structures. In this way, each molecule is represented as a graph $\mathcal{G}$ that can be inputted into the model for property prediction. Finally, the prediction result $Y$ is obtained.

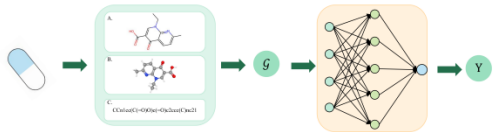

Figure 5: different Representation of compounds

## Experiments

In this section, experimental validation on the benchmark dataset is performed to investigate the rationality of ADMEOOD. The typical experimental results and their corresponding analysis are provided. More results and experimental details are provided in Appendix C.

### Performance of baseline algorithms

In the benchmarking of drug property prediction, experiments on 24 datasets with 4 baseline methods are conducted. For each dataset, the same GIN backbone is used for all baseline methods for fair comparisons.

To validate the performance gap, experiments are conducted using ERM and the results are summarized in Table 1. The evaluation metric used in the results is AUROC, which indicates the classifier's ability to distinguish between classes. The higher the AUC score, the better the model's performance in classification. From the table, it can be seen that the model's performance significantly decreases from in-distribution (IID) to out-of-distribution (OOD). In the assay domain, the AUC scores of ERM in the IID validation sets of Noise Shift and CCD are 91.97% and 93.60% respectively, while in the OOD validation sets of Noise

Shift and CCD, the AUC scores are 83.59% and 59.89% respectively. For the scaffold domain, ERM achieves AUC scores of 90.54% and 99.98% on the ID validation set of Noise Shift and the ID validation set of CCD, respectively. Additionally, it achieves scores of 86.32% and 73.08% on the OOD validation set of Noise Shift and the OOD validation set of CCD, respectively. Under two different data shift methods, the performance of ERM decreased by AUC scores of 8.38%, 33.71%, 4.22%, and 26.9% separately. This shows that these splitting methods indeed have significantly better and more stable effects compared to conventional random splitting.

| Method | Noise Shift | | CCD | |
|---|---|---|---|---|
| | IID(AUC) | OOD | IID | OOD |
| Assay | 91.97 | 83.59 | 93.60 | 59.89 |
| Scaffold | 90.54 | 86.32 | 99.98 | 73.08 |

Table 1: Results of ERM on datasets with different domain shift: NS-ec50-assay, CCD-ec50-assay, NS-ec50-scaffold, and CCD-ec50-scaffold.

The performance gap in different domain partitions is shown in Figure 6. Gap values are computed on the validation set and averaged across all measurement types. As shown from the results, in Noise Shift, the scaffold domain cause significant performance degradation because molecules from different backbones often have different properties and noise may mask or interfere with features that are related to specific properties or activities. When it comes to CCD, the assay domain can lead to a drop in performance, as various labs carrying out the same experiment under different circumstances may produce varying outcomes. This could have a major effect on the labels.

### Quantitative comparison and analysis

The results of the 4 baseline models for two methods are shown in Table 2. Specifically, in a dataset from the CCD-assay domain, IRM outperforms other methods in multiple random environments. This demonstrates that the invariance objective of IRM can effectively mitigate performance degradation in this particular case. In order to demonstrate the learning invariance and robustness of the model, the results are averaged from the optimal values obtained in different environments with different random seeds. In the dataset of CCD-scaffold, Mixup performs excellently in different environments, especially in the case of measuring EC50, where its AUC reaches 76.11%. According to the results of Noise Shift, ERM consistently performs better than others. This suggests that current domain generalization methods are not effective in addressing OOD problem caused by noise. In the scaffold domain, IRM and DeepCoral achieve AUC scores of 79.97% and 79.65% respectively on the OOD validation set, while the AUC score for the ERM baseline is 83.59%. In summary, although these models have made some improvements, other flaws in algorithm design have caused performance bottlenecks. This indicates the need to develop better algorithms to solve OOD problem in the field of drug property prediction.

| Domain | Assay | | Scaffold | | |
|---|---|---|---|---|---|
| Algos | Val (IID) | Val (OOD) | Val (IID) | Val (OOD) | |
| ERM | 93.60(0.12) | 59.89(2.60) | 99.98(0.01) | 73.08(3.79) | CCD |
| IRM | 83.42(1.21) | **65.68(1.90)** | 87.19(1.78) | 71.43(2.18) | |
| DeepCoral | 88.01(0.82) | 62.36(1.83) | 88.27(1.94) | 72.01(3.22) | |
| Mixup | 93.11(0.19) | 58.44(1.83) | 99.57(0.03) | **76.11(2.64)** | |
| ERM | 91.97(0.07) | **83.59(0.15)** | 90.54(0.13) | **86.32(0.34)** | Noise Shift |
| IRM | 83.57(0.11) | 79.97(0.86) | 89.77(1.44) | 83.23(0.70) | |
| DeepCoral | 84.04(0.14) | 79.65(1.20) | 91.75(1.25) | 83.71(0.13) | |
| Mixup | 91.20(0.23) | 82.49(0.07) | 99.61(0.01) | 84.07(0.30) | |

Table 2: The baseline results of the four OOD algorithms on datasets from different domains in the two data shifting approaches are as follows. The results are reported in AUC score.

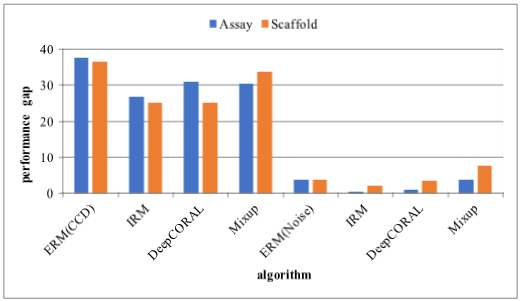

Figure 6: Performance gap on AUC of different domain splits with different data shift methods and OOD algorithms on the ADMEOOD. Gap values are computed on the OOD validation set. Noise Shift on the left, CCD on the right.

## Conclusion

In this work, a systematic OOD dataset curator and benchmark specifically designed for drug property prediction called ADMEOOD is proposed. It includes two kinds of OOD data shifts: Noise Shift and CCD. Subsequently, instance testing is conducted to compare four contemporaries classic OOD generalization algorithms. The results of the study reveal significant performance differences between IID and OOD data. Existing OOD algorithms can improve generalization ability, but their effectiveness is not significantly better than the baseline ERM method. An algorithm might improve performance on one type of shift, but not both. With these observation results, future OOD methods can focus on addressing Noise Shift and CCD issues to improve specific generalization capabilities. It can be achieved by using a more appropriate model architecture or adopting causal inference strategies. In addition, different factors such as experimental accuracy, measurement types, data sources, etc., will all have an impact on the performance of the model. By further denoising techniques, it is possible to improve the quality of the data, thereby enabling the training of more precise drug property prediction models. ADMEOOD is a growing project, and we hope that in the future it will include a greater number and variety of datasets and domain selections. In addition, the benchmark will further develop OOD generalization algorithms to enhance the performance of the model.

## Acknowledgment

This work was supported in part by the National

# References

Alexov, E.; Mehler, E. L.; Baker, N. A.; Schwab, D. J.; Huang, Y.; Milletti, F.; Nielsen, J.; Farrell, D.; Carstensen, T.; Olsson, M.; Payne, G. F.; Warwicker, J.; Williams, S.; and Word, J. M. 2011. Progress in the prediction of p$K_a$ values in proteins. *Proteins*, 79(12): 3260-3275.

Arjovsky, M.; Bottou, L.; Gulrajani, I.; and Lopez-Paz, D. 2019. Invariant risk minimization. *arXiv preprint arXiv:1907.02893*.

Benfenati, E.; Gini, G.; Piclin, N.; Roncaglioni, A.; and Vari, M. R. 2003. Predicting logP of pesticides using different software. *Chemosphere*, 53(9): 1155-1164.

Burggraaff, L.; Van Vlijmen, H.; IJzerman, A. P.; and Van Westen, G. J. P. 2020. Quantitative prediction of selectivity between the A1 and A2A adenosine receptors. *Journal of Cheminformatics*, 12(1): 1-16.

Cortés-Ciriano, I.; and Bender, A. 2016. How consistent are publicly reported cytotoxicity data? Large-scale statistical analysis of the concordance of public independent cytotoxicity measurements. *ChemMedChem*, 11(1): 57-71.

Cruciani, G.; Milletti, F.; Storchi, L.; Sforna, G.; and Goracci, L. 2009. In silico p$K_a$ Prediction and ADME Profiling. *Chemistry & Biodiversity*, 6(11): 1812-1821.

GaninYaroslav.; UstinovaEvgeniya.; AjakanHana.; GermainPascal.; LarochelleHugo.; LavioletteFrançois.; MarchandMario.; and LempitskyVictor. 2016. Domain-adversarial training of neural networks. *Journal of Machine Learning Research*, 17(1): 2096-2030.

Gaulton, A.; Bellis, L. J.; Bento, A. P.; Chambers, J. K.; Davies, M. G.; Hersey, A.; Light, Y.; McGlinchey, S.; Michalovich, D.; Al-Lazikani, B.; and Overington, J. P. 2011. ChEMBL: a large-scale bioactivity database for drug discovery. *Nucleic Acids Research*, 40(D1): D1100-D1107.

Gertrudes, J. C.; Maltarollo, V. G.; Silva, R. A.; Oliveira, P. R.; Honorio, K. M.; and Da Silva, A. B. F. 2012. Machine learning techniques and drug design. *Current medicinal chemistry*, 19(25): 4289-4297.

Greene, N. 2002. Computer systems for the prediction of toxicity: an update. *Advanced Drug Delivery Reviews*, 54(3): 417-431.

Gui, S.; Li, X.; Wang, L.; and Ji, S. 2022. Good: A graph out-of-distribution benchmark. *Advances in Neural In-*

formation Processing Systems, 35: 2059-2073.

Ji, Y.; Zhang, L.; Wu, J.; Wu, B.; Li, L.; Huang, L.; Xu, T.; Rong, Y.; Ren, J.; Xue, D.; Lai, H.; Liu, W.; Huang, J.; Zhou, S.; Luo, P.; Zhao, P.; and Bian, Y. 2023. DrugOOD: Out-of-Distribution Dataset Curator and Benchmark for AI-Aided Drug Discovery-a Focus on Affinity Prediction Problems with Noise Annotations. In *Proceedings of the AAAI Conference on Artificial Intelligence,* volume 37, 8023-8031.

Kennedy, T. 1997. Managing the drug discovery/development interface. *Drug discovery today*, 2(10): 436-444.

Kipf, T. N.; and Welling, M. 2016. Semi-supervised classification with graph convolutional networks. *arXiv preprint arXiv:1609.02907*.

Koh, P. W.; Sagawa, S.; Marklund, H.; Xie, S. M.; Zhang, M.; Balsubramani, A.; Hu, W.; Yasunaga, M.; Phillips, R. L.; Gao, I.; Lee, T.; David, E.; Stavness, I.; Guo, W.; Earnshaw, B.; Haque, I.; Beery, S. M.; Leskovec, J.; Kundaje, A.; Pierson, E.; Levine, S.; Finn, C.; and Liang, P. 2021. WILDS: A Benchmark of in-the-Wild Distribution Shifts. In Meila, M.; and Zhang, T., eds., *Proceedings of the 38th International Conference on Machine Learning, volume 139 of Proceedings of Machine Learning Research*, 5637-5664. PMLR.

Kola, I.; and Landis, J. 2004. Can the pharmaceutical industry reduce attrition rates? *Nature reviews Drug discovery*, 3(8): 711-716.

Kramer, C.; Kalliokoski, T.; Gedeck, P.; and Vulpetti, A. 2012. The experimental uncertainty of heterogeneous public K i data. *Journal of medicinal chemistry*, 55(11): 5165-5173.

Kuhn, M.; Letunic, I.; Jensen, L. J.; and Bork, P. 2016. The SIDER database of drugs and side effects. *Nucleic acids research*, 44(D1): D1075-D1079.

Landrum, G. 2013. Rdkit documentation. *Release*, 1(1-79): 4.

Lu, C.; Liu, Q.; Wang, C.; Huang, Z.; Lin, P.; and He, L. 2019. Molecular property prediction: A multilevel quantum interactions modeling perspective. In *Proceedings of the AAAI Conference on Artificial Intelligence,* volume 33, 1052-1060.

Mayr, A.; Klambauer, G.; Unterthiner, T.; and Hochreiter, S. 2016. DeepTox: toxicity prediction using deep learning. *Frontiers in Environmental Science*, 3: 80.

O'Boyle, N. M.; Banck, M.; James, C. A.; Morley, C.; Vandermeersch, T.; and Hutchison, G. R. 2011. Open Babel: An open chemical toolbox. *Journal of cheminformatics*, 3(1): 1-14.

Riedel, S.; Yao, L.; McCallum, A.; and Marlin, B. M. 2013. Relation extraction with matrix factorization and universal schemas. In *Proceedings of the 2013 Conference of the North American Chapter of the Association for Computational Linguistics: Human Language Tech-* *nologies*.

Rupp, M.; Körner, R.; and Tetko, I. V. 2011. Predicting the PKA of small molecules. *Combinatorial Chemistry & High Throughput Screening*, 14(5): 307-327.

Sheridan, R. P.; Wang, W. M.; Liaw, A.; Ma, J.; and Gifford, E. M. 2016. Extreme gradient boosting as a method for quantitative structure-activity relationships. *Journal of chemical information and modeling*, 56(12): 2353-2360.

Sun, B.; and Saenko, K. 2016. Deep coral: Correlation alignment for deep domain adaptation. In *Computer Vision–ECCV 2016 Workshops,* volume 9915, 443-450.

Tan, Y.; Dai, L.; Huang, W.; Guo, Y.; Zheng, S.; Lei, J.; Chen, H.; and Yang, Y. 2022. DRLinker: Deep Reinforcement Learning for Optimization in Fragment Linking Design. *Journal of Chemical Information and Modeling*, 62(23): 5907-5917.

Velickovic, P.; Cucurull, G.; Casanova, A.; Romero, A.; Lio, P.; and Bengio, Y. 2017. Graph attention networks. *Stat.ML*, 1050(20): 10-48550.

Wang, S.; Guo, Y.; Wang, Y.; Sun, H.; and Huang, J. 2019. Smiles-bert: large scale unsupervised pre-training for molecular property prediction. In *Proceedings of the 10th ACM International Conference on Bioinformatics, Computational Biology and Health Informatics*.

Weininger, D. 1988. SMILES, a chemical language and information system. 1. Introduction to methodology and encoding rules. *Journal of Chemical Information and Computer Sciences*, 28(1): 31-36.

Wieder, O.; Kohlbacher, S.; Kuenemann, M.; Garon, A.; Ducrot, P.; Seidel, T.; and Langer, T. 2020. A compact review of molecular property prediction with graph neural networks. *Drug Discovery Today: Technologies*, 37: 1-12.

Witham, S.; Talley, K.; Wang, L.; Zhang, Z.; Sarkar, S. J.; Gao, D.; Yang, W.; and Alexov, E. 2011. Developing hybrid approaches to predict p$K_a$ values of ionizable groups. *Proteins*, 79(12): 3389-3399.

Wu, Z.; Ramsundar, B.; Feinberg, E. N.; Gomes, J.; Geniesse, C.; Pappu, A. S.; Leswing, K.; and Pande, V. S. 2018. MoleculeNet: a benchmark for molecular machine learning. Chemical science, 9(2): 513-530.

Xing, L.; and Glen, R. C. 2002. Novel Methods for the Prediction of logP, p$K_a$, and logD. *Journal of chemical information and computer sciences*, 42(4): 796-805.

Xu, K.; Hu, W.; Leskovec, J.; and Jegelka, S. 2018. How powerful are graph neural networks? *arXiv preprint arXiv:1810.00826*.

Xu, M.; Zhang, J.; Ni, B.; Li, T.; Wang, C.; Tian, Q.; and Zhang, W. 2020. Adversarial Domain Adaptation with Domain Mixup. In *The Thirty-Fourth AAAI Conference on Artificial Intelligence*, 6502-6509. AAAI Press.

Zhang, H.; Cisse, M.; Dauphin, Y. N.; and Lopez-Paz, D. 2017. mixup: Beyond empirical risk minimization. *arXiv preprint arXiv:1710.09412*.

Zhou, F.; Jiang, Z.; Shui, C.; Wang, B.; and Chaib-draa, B. 2021. Domain Generalization via Optimal Transport with Metric Similarity Learning. *Neurocomputing*, 456(C): 469-480.

Zhu, J.; Xia, Y.; Qin, T.; Zhou, W.; Li, H.; and Liu, T. Y. 2021. Dual-view molecule pre-training. *arXiv preprint arXiv:2106.10234*.